\pdfoutput=1

\documentclass[11pt]{article}

\usepackage[]{emnlp2023}

\usepackage{times}
\usepackage{latexsym}
\usepackage{enumitem}
\usepackage{booktabs}
\usepackage{graphicx}
\usepackage{amsmath}
\usepackage{amssymb}%
\usepackage{verbatim}
\usepackage[belowskip=-15pt,aboveskip=10pt]{caption}
\usepackage{pifont}
\usepackage{xcolor}

\usepackage[T1]{fontenc}

\usepackage[utf8]{inputenc}
\usepackage{graphics}
\usepackage{microtype}

\usepackage{inconsolata}

\title{Help Me Identify: \\ Is an LLM+VQA System All We Need to Identify Visual Concepts?}

\author{Shailaja Keyur Sampat  \and Maitreya Patel \and Yezhou Yang \and Chitta Baral 
\\ \texttt{\{ssampa17, mpatel57, yz.yang, chitta\}@asu.edu} \\ 
         Arizona State University}

\begin{document}

\maketitle

\begin{abstract}
An ability to learn about new objects from a small amount of visual data and produce convincing linguistic justification about the presence/absence of certain concepts (that collectively compose the object) in novel scenarios is an important characteristic of human cognition. This is possible due to abstraction of attributes/properties that an object is composed of e.g. an object `bird' can be identified by the presence of a beak, feathers, legs, wings, etc. Inspired by this aspect of human reasoning, in this work, we present a zero-shot framework for fine-grained visual concept learning by leveraging large language model and Visual Question Answering (VQA) system\footnote{Code is available at \url{https://github.com/shailaja183/ObjectConceptLearning.git}}. Specifically, we prompt GPT-3 to obtain a rich linguistic description of visual objects in the dataset. We convert the obtained concept descriptions into a set of binary questions. We pose these questions along with the query image to a VQA system and aggregate the answers to determine the presence or absence of an object in the test images. Our experiments demonstrate comparable performance with existing zero-shot visual classification methods and few-shot concept learning approaches, without substantial computational overhead, yet being fully explainable from the reasoning perspective.

\end{abstract}

\section{Introduction}

Imagine a toddler looking at a new picture book. Suddenly it points to an illustration and shouts `chair'. Although the chairs that the kid has observed in the house are different (in designs and colors) from the ones shown in the picture book, how are they able to correctly identify the object chair? The answer is- abstraction and categorization, which are two fundamental elements of human cognition \cite{reinert2021mouse}. 

In other words, objects are the building blocks of our understanding of the world. In our day-to-day lives, we constantly encounter diverse objects in our surroundings- by looking at images, reading textual descriptions, interacting with objects, or conversing with other humans. Our brain continuously categorizes the perceived information and develops mental representations of various objects by abstracting their properties \cite{murphy2010categories}. For the above example, a mental representation of the object `chair' is formed using attributes such as having a backrest, legs, a seat, made of wood or plastic, etc. This allows the toddler to effortlessly recognize novel kinds of chairs. 

The most common approach for distinguishing objects in Vision-Language (V\&L) models is through supervised training over a set of labeled images. Despite remarkable advances in the field, the best existing systems still rely on a large number of annotated training examples. On the other hand, humans (including children) typically need only a handful of examples to robustly recognize the object and make meaningful generalizations \cite{landau1988importance}.

Another drawback of existing V\&L models for object recognition is their black-box nature. These models rely on the similarity of visual features (obtained from the images) and possible class labels to distinguish between instances. As a result, the model's reasoning process behind a particular prediction is not explainable. Contrary to that, if a person is asked to identify a particular object, they would pay closer attention to the attributes (concepts) that an object is likely to have. Such a modular reasoning process allows humans to explain their decision linguistically. For example, because a particular bird has a red body, and pointed crest (along with other features), a person is likely to identify the bird as `cardinal'.

Inspired by this aspect of human reasoning, we propose a training-free concept learning strategy that is driven by large language models (LLMs). Our hypothesis here is that LLMs trained on massive text corpora possess a vast amount of knowledge about various concepts. Such models can be queried to produce useful textual descriptions about visual appearance of an object. It can be followed by Visual Question Answering (VQA) systems to verify the presence/absence of LLM-generated concept descriptions in the unseen images. This will alleviate the reliance on visual recognition systems that are expensive in nature (from supervised data collection and training viewpoints) yet crucial for many V\&L downstream tasks. 

\paragraph{Contributions:}
\begin{itemize}[noitemsep,topsep=0pt]
    \item We present a lightweight, interpretable framework for zero-shot visual concept learning by combining an LLM with VQA systems. 
    \item We demonstrate the effectiveness of our approach over several fine-grained visual classification benchmarks in comparison with state-of-the-art methods.
    \item We perform relevant ablations and provide a detailed analysis of the strengths and weaknesses of our proposed approach with respect to the CUB dataset \cite{wah2011caltech}.
\end{itemize}

\section{Related Work}

In this section, we summarize existing works along three research directions (within vision-language) that are closely related to our work in this paper. We highlight similarities and distinctions of our proposed method with advancements in these areas.   

\subsection{Interpretable Image Classification}

Image classification research soared from hand-crafted features \cite{lowe1999object, dalal2005histograms} to deep learning models \cite{simonyan2014very, he2016deep}, gaining accuracy at the cost of interpretability \cite{kamakshi2023explainable}. To address this opaqueness of deep learning models in their decision-making, two broad sets of methods have been proposed- Posthoc and Antehoc. Posthoc methods aim to explain the behavior of an AI system post-training, through model probing (by observing the model's decision over diverse input-output pairs). Whereas, antehoc approaches aim to develop models that are interpretable by design, either through modification of existing architectures or by the addition of novel components to support explainability. 

Most commonly used posthoc approaches include (i) saliency or class activation maps- where image regions are highlighted by color-coding based on their importance \cite{selvaraju2017grad, chattopadhay2018grad}; (ii) counterfactual explanations- where alternative visual scenarios are generated to explain the behavior of the model \cite{goyal2019counterfactual, lang2021explaining, wang2020scout}, and (iii) concept-based approaches- where explanations (in the form of vector representations) can be mapped to resemble human-like thinking in recognition using colors, body parts, textures etc. \cite{koh2020concept, chen2019looks}. Contrary to that, (i) generation of textual descriptions about how an image classifier makes its predictions \cite{hendricks2016generating, hendricks2018grounding}; and (ii) neuro-symbolic models- where training process is guided by human-curated knowledge base or ontology \cite{marino2016more, alirezaie2018symbolic} fall under antehoc methods.  

Our proposed approach in this paper is relevant to both posthoc and antehoc explanation methods. The use of LLM as a knowledge base and leveraging textual description of visual objects as an explanation mechanism is relevant to the antehoc approach. On the other hand, the use of pre-trained VQA models and probing its responses with respect to the given image+question (input) and answer (output), aligns with the posthoc approach.

\subsection{Visual Concept Learning Methods}
\label{sec:vclmlit}
Existing methods that tackle visual concept learning can be divided into two major kinds: meta-learning-based methods and foundation model-based methods. \citet{mei2021falcon, cao2020concept} are two works that leverage meta-learning techniques for concept recognition. In particular, \citet{mei2021falcon} use $<$image, caption, supplemental sentences$>$ triples to learn box embedding space that represents visual objects in a dataset. At the test time, the learned box embeddings are used to answer questions about unseen images. On the other hand, COMET \cite{cao2020concept} learns concept embeddings using independent concept learners and compares them to concept prototypes. Later, it effectively aggregates the information across concept dimensions, by assigning concept importance scores to each dimension. 

The second kind of methods use GPT-3 \cite{brown2020language} as a knowledge base to tackle visual concept learning task. LaBo \cite{yang2023language} extended the idea of concept bottleneck models \cite{koh2020concept} with LLM-extracted concepts which supports few-shot evaluation. \citet{menon2022visual, pratt2022does, yan2023learning} generate concept descriptions with LLMs and use them for knowledge-aware zero-shot image-text matching over CLIP \cite{radford2021learning} to improve concept recognition. While the focus of \citet{pratt2022does} (referred to as CuPL) was on automatic selection of the prompt to query the LLM, \citet{yan2023learning} (referred to as CDA) distinguished themselves by identifying a small concise set of relevant attributes by avoiding noisy LLM generations and reduced computation.  

We share two-fold similarity with all aforementioned approaches: first, zero-shot/few-shot evaluation of visual concept learning, and second, interpretable model development. Our approach falls under foundation models-based visual concept learning methods. We rely on a fixed hand-crafted prompt like \citet{menon2022visual, yang2023language} to obtain visual descriptors for each object. However, we do not consider object labels during the visual recognition step like \citet{menon2022visual}. Unlike all foundation model-based methods that rely on CLIP for image-attribute matching, in this work, we explore VQA models to verify the presence or absence of attributes in the given image. Our approach yields better or comparable performance over existing methods considering only three concept descriptions per object (i.e. m=3), which is significantly lower than \citet{yan2023learning} (i.e. m$\geq$16). 

\subsection{Leveraging Foundation Models for Vision-Language (V\&L) Tasks}

Recently, large language models (LLMs) have gained popularity with their impressive capabilities in language generation, in-context learning, and reasoning \cite{brown2020language, touvron2023llama, penedo2023refinedweb}. These models possess a vast amount of world knowledge \cite{petroni2019language, roberts2020much, hendrycks2020measuring} due to training on massive text corpora. Inspired by this, several works investigated if LLMs can provide useful biases to improve downstream tasks in the vision-language domain.

\citet{bianco2023improving} proposed a method to enhance the informativeness of model-generated image captions using LLMs. \citet{song2022llm} utilized LLM as a high-level planner for embodied instruction following in a visually-perceived environment. In Img2LLM system \cite{guo2023images}, synthetic QA pairs were generated as in-context exemplars for a given image to improve on the zero-shot VQA task. All the above papers demonstrated that the knowledge from LLMs and the abilities of visual reasoning models combined can achieve comparable results on existing V\&L tasks without requiring explicit pre-training or fine-tuning. Further, MiniGPT-4 \cite{zhu2023minigpt} demonstrated that it is even possible to achieve multi-modal language model-like capabilities by combining a frozen visual encoder with a frozen language model using one projection layer. This model was shown to be able to do unconventional V\&L tasks such as writing stories and poems based on images, generating recipes from photos of food, etc. 

Our work in this paper is closest to \citet{guo2023images}, as both works leverage LLMs for zero-shot VQA by decomposing the task into multiple sub-questions. However, their goal is to connect  immediate contents of the image with broader contextual information using LLMs. In contrast, we are interested in obtaining fine-grained visual descriptions of various objects to recognize them in an interpretable manner using the VQA task as a proxy.  

\section{Task and Evaluation}
\label{sec:llmvqatask}

\subsection{Definition: Visual Concept Learning} 
In cognitive psychology research, `concepts' are defined as a set of mental categories that aid the grouping of various objects, events, or ideas based on their common relevant features \cite{murphy2004big}. Therefore, `concept learning' is defined as- the search for and listing of attributes that can be used to distinguish exemplars from non-exemplars of various categories \cite{bruner1986study}. In artificial intelligence terminology, the above definition can be translated as- identifying a set of features that construct a particular concept from a set of positive and negative training examples related to that concept \cite{learning1997tom}.  Concept learning about visual appearances (i.e. how a particular concept looks like) is referred to as visual concept learning. Following \cite{bruner1986study}, in this work, we consider  attributes of the object visible in the image (such as colors, shapes, size, body parts, textures etc.) as visual concepts that can be compositely used to identify the object category and distinguish it from other objects.

\begin{figure}[ht!]
\includegraphics[width=\linewidth]{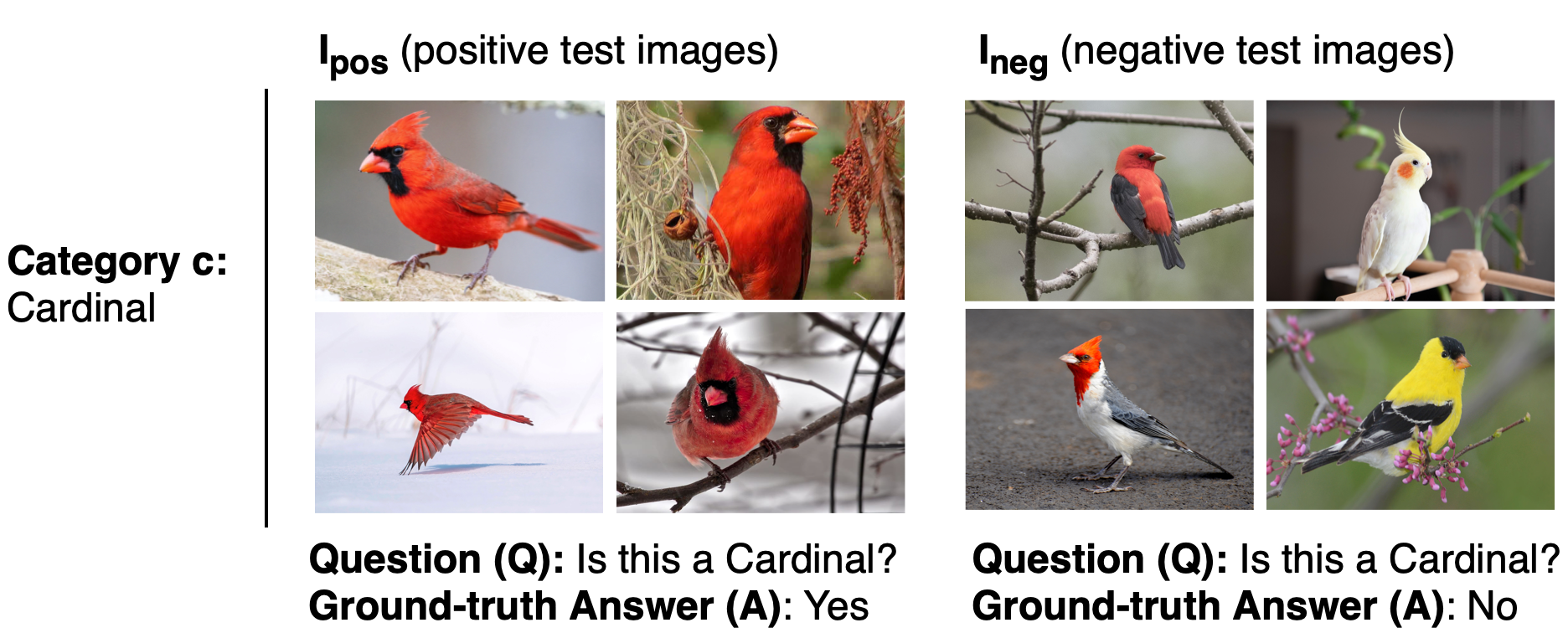}
\caption{Zero-shot VQA task considered in this paper demonstrated using `cardinal' object category from CUB \cite{wah2011caltech}:  provided an image and a question as input, the model has to predict `Yes'/`No' answer depending on the presence or absence of the objects.}
\label{fig:taskexample}
\end{figure}

\subsection{Task: Zero-shot  Visual Question Answering} 

 In this paper, we exclusively investigate zero-shot visual question answering (VQA) task for visual concept learning (as shown in Figure \ref{fig:taskexample}). For each object category $c$, a set of testing examples are provided. Each example consists of an image and a question pair, for which a model should predict an answer. Images are considered positive (I$_{pos}$) or negative (I$_{neg}$) depending on whether the category $c$ present or not respectively. All images in I$_{pos}$ would have the category $c$ present, but images may demonstrate a broad range of view-variations (different pose/action, view angle, crop factor or relative position of $c$ in the frame, etc.). Whereas, images in the negative set I$_{neg}$ are about different object category $c^{\prime}$ such that $c^{\prime}\neq c$. For each image in I$_{pos}$ $\cup$ I$_{neg}$ for a category c, a binary question Q of the form `Is there a $<$c$>$?' is posed. For all images in I$_{pos}$ set, the answer to the question Q would be `Yes' (indicating the presence of object category c). Whereas for the images in set I$_{neg}$, the answer would be `No' (indicating the absence of object category c). 

\subsection{Notations Used for Experimental Evaluation} 

There are three important  parameters that distinguish various methods for visual concept learning using LLMs. The parameter values/ranges reported in this paper are determined by respective authors (adapted from original manuscripts) based on their architecture design or empirical observations. 

\begin{itemize}[noitemsep, leftmargin=*]
   \item \textbf{m} refers to the number of concepts generated to identify a particular object category c in the image and distinguish it from all other object categories $c^{\prime}\neq c$.
\item  \textbf{n} refers to n-shot training or use of n labeled training examples for the preparation of a model; n=0 suggests that a given model is test-only or zero-shot (no training phase incorporated).
\item  \textbf{p} indicates whether the underlying method uses the fixed or variable prompt; Fixed prompt refers to the use of identical prompt for all object categories to obtain concept description using the LLM. Whereas variable prompt means a custom prompt is generated for every object category by leveraging prior knowledge (such as from external object taxonomy or additional information visible in the image).
\end{itemize}


\section{Proposed Method: Leveraging LLMs+VQA Systems for Interpretable Concept Learning}
\label{sec:llmvqa}

\begin{figure*}
\centering
\includegraphics[width=0.8\linewidth]
{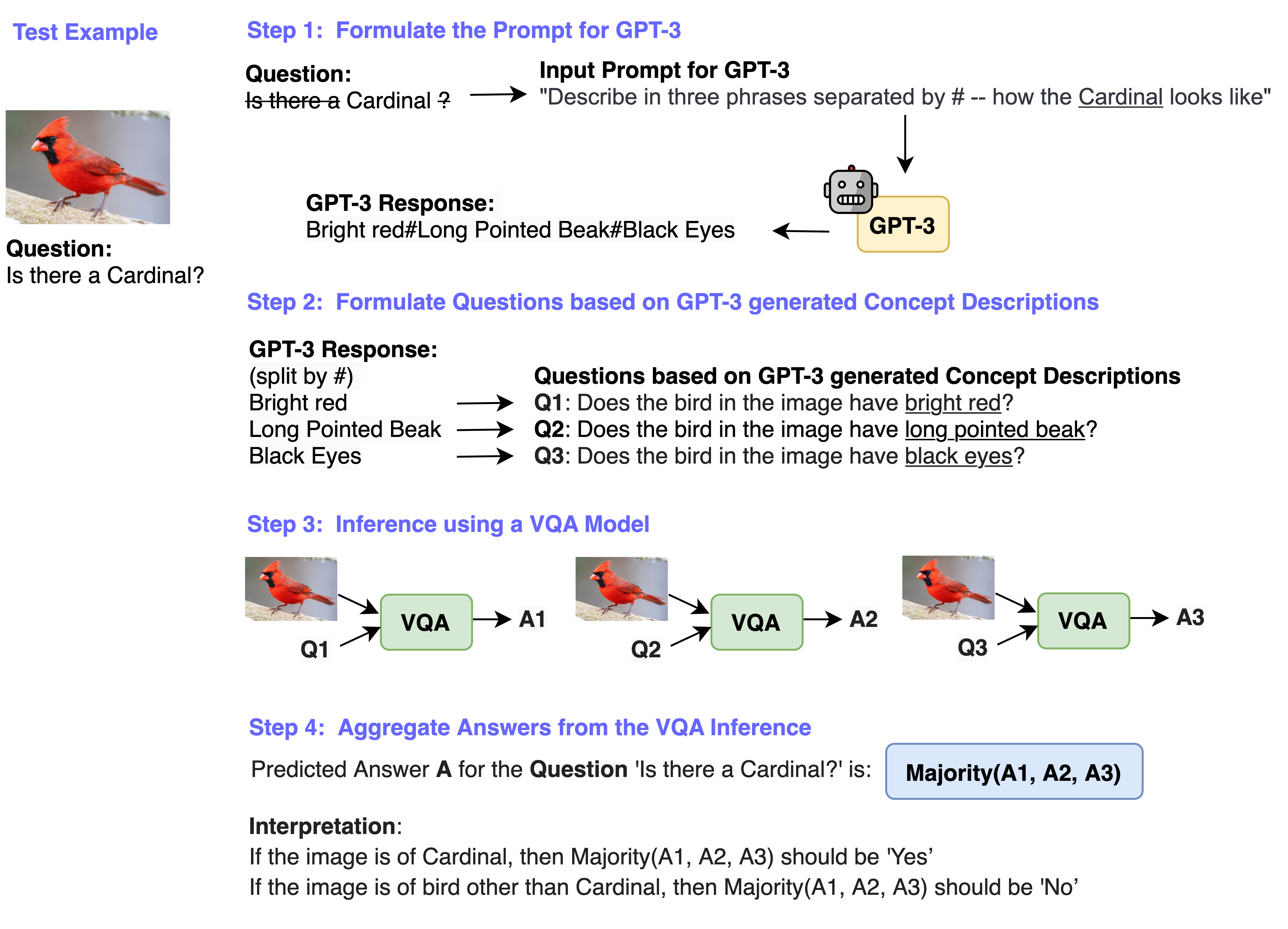}
\caption{Overview of our proposed zero-shot method that uses LLM+VQA for concept learning: There are four key steps- (i) given an object category that needs to be verified in the image, we first query GPT-3 using a predefined prompt to obtain the concept descriptions for the object, (ii) concept descriptions returned by GPT-3 are turned into a set of binary meta-questions, (iii) test image along with each  meta-question is posed to a VQA system, (iv) aggregate answers of all meta-questions to determine presence or absence of an object category in the image.}
\label{fig:cppl}
\end{figure*}

In this section, a zero-shot approach is described, which leverages knowledge of various concepts in large language model (GPT-3) and fine-grained visual understanding capabilities of visual question answering (VQA) systems. Figure \ref{fig:cppl} visually demonstrates the proposed approach- which can be divided into four key steps; 

\paragraph{Step 1: Obtaining Concept Descriptors using GPT-3} 

Consider a test case from the CUB \cite{wah2011caltech}- fine-grained bird image classification dataset shown in Figure \ref{fig:cppl}. The objective is to verify whether or not `cardinal' is present in the image, as per the question. The first step is to query GPT-3 \cite{brown2020language} to retrieve concept descriptions. We particularly use a template-based prompt of form `Describe in m phrases separated by \# -- how the $<$object\_category$>$ looks like'. Where m is an odd integer and $<$object\_category$>$ is the object category that is to be verified in the image. More generally, as a result of this step, GPT-3 would generate m concept descriptions \{v$_1$, v$_2$, .., v$_m$\} separated by \#. For the example at hand, using m=3 and substituting $<$object\_category$>$ as `cardinal', GPT-3 generates the response `Bright red\#Long Pointed Beak\#Black Eyes'. These concept descriptions capture GPT-3's understanding about visual attributes of a cardinal.

\begin{figure}[ht!]
\includegraphics[width=\linewidth]{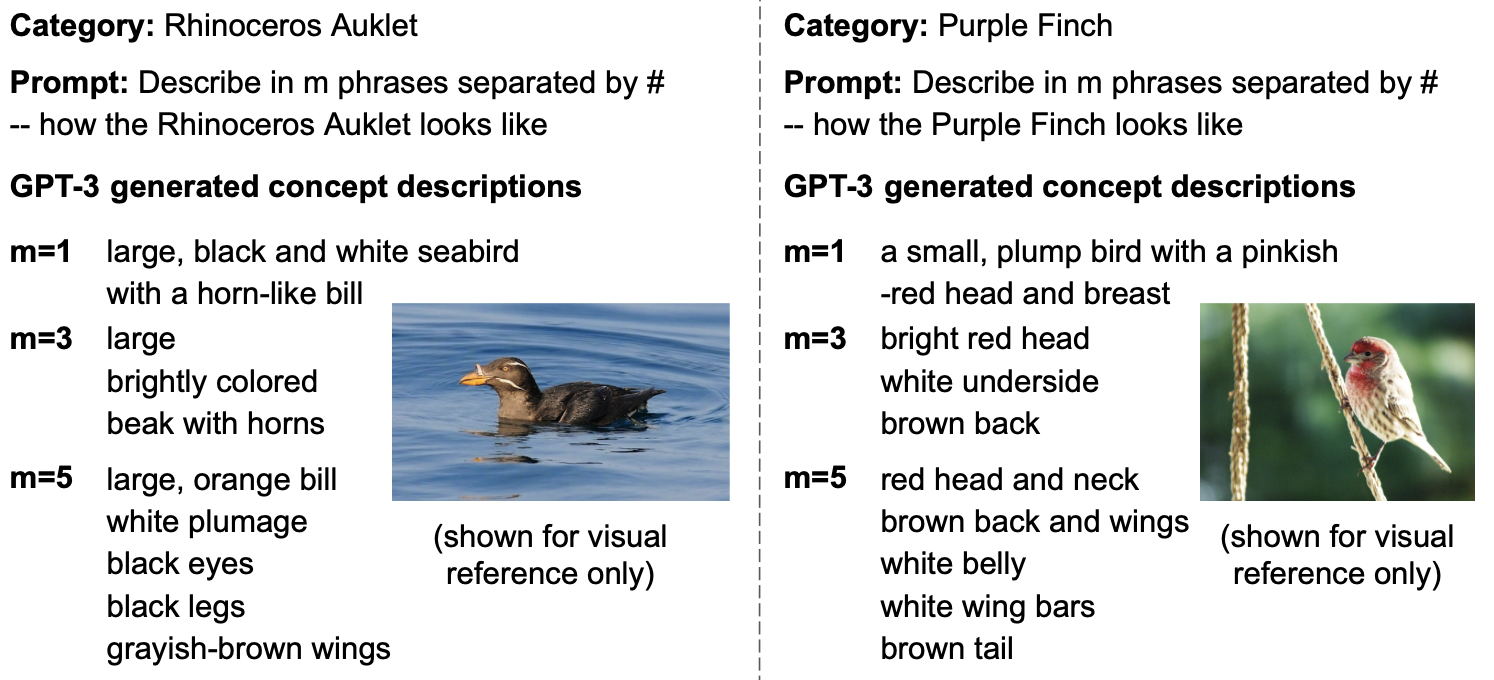}
\caption{Two categories from the CUB dataset and their fine-grained concept descriptions generated by GPT-3 for m=\{1,3,5\}.}
\label{fig:cubmvar}
\end{figure}

Our system works with odd values of m as we use the majority as the aggregation mechanism in the later step. In this paper, we experiment with m=\{1,3,5\} and report their results. Figure \ref{fig:cubmvar} demonstrates two examples from the CUB \cite{wah2011caltech} dataset when GPT-3 is prompted with different values of m. It can be observed that when m=1, the generated concept descriptions are longer and typically includes a description of more than one body parts and its visual attributes. With m=3 and m=5, the generated phrases are short, concise, and discriminative from each other. A more detailed analysis of GPT-3 generated concept descriptions is provided in Section \ref{sec:res}. 

\paragraph{Step 2: Generating Binary Meta-questions}

The concept descriptions generated using GPT-3 (as described in the previous step) are in the form of phrases. For this task, we are interested in investigating whether or not existing visual question answering (VQA) systems can verify the fine-grained visual descriptions generated by GPT-3 in the images. The VQA systems take questions as linguistic input. Therefore, using a simple template `Does the bird in the image have $<$v$_i$$>$?' (where $i$ ranges from 1 to m), we convert GPT-3 generated phrases into a set of m binary questions. This we refer to as meta-questions corresponding to the concepts identified by GPT-3 for a given category of objects. For the cardinal example,  three questions are formed: (1) Does the bird in the image have Bright red?, (2) Does the bird in the image have Long Pointed Beak?, and (3) Does the bird in the image have Black Eyes?.

\paragraph{Step 3: Inference using VQA Models}

As mentioned earlier, our proposed method is primarily designed for zero-shot concept learning. We take the best off-the-shelf single-model architectures that support VQA tasks. Particularly, we consider BLIP \citep{li2022blip}, GIT \citep{wang2022git} and ViLT \citep{kim2021vilt}. For each architecture, we take a pre-trained version of the model fine-tuned on the VQA dataset \citep{antol2015vqa}. We pose these models with the test image and meta-questions corresponding to the category (obtained from Step 2). In this case, our hypothesis is that- since these models are previously fine-tuned on the VQA dataset, they would be able to recognize different body parts of the bird and associated attributes (such as color, shape, size, texture).  

Particularly, BLIP \citep{li2022blip} is a vision-language pre-training framework that effectively utilizes a synthetic caption generation along with a filter to identify noisy captions. It uses a multi-modal mixture of encoder-decoder architecture over large-scale noise-filtered image-caption data. It can transfer its learning well to both vision-language understanding and generation tasks in comparison with its predecessors. GIT \citep{wang2022git} is a generative image-to-text transformer, which is simple network architecture (consisting of only one image encoder and one text decoder) that achieves strong performance with data-scaling. It can be used to perform a variety of vision-language tasks including visual question answering. Whereas, ViLT \citep{kim2021vilt} is a convolution-free pre-training approach that eliminates the need for obtaining object detection, object tags, OCR (optical character recognition), and region features from the image, which is faster, more parameter efficient yet demonstrates strong performance on downstream vision-language tasks.

\paragraph{Step 4: VQA Answers Aggregation}

Since each test image has m meta-questions corresponding, there are m binary (Yes/No) answers are generated in step 3. The final step in our approach is to aggregate the answers to conclude the presence or absence of the object category. For m=1, there is only one binary meta-question generated. For a positive example, the affirmative answer to the meta-question would determine the presence whereas for a negative example, the non-affirmative answer would indicate the absence of the concept. For m>1, a set of m answers will be generated by the VQA system therefore we use majority as an aggregate measure for decision making. For this reason, we suggest taking odd values of m when working with our proposed model.    

For m=3, if at least two answers among the three meta-questions for the positive and negative test cases are `Yes' and `No' respectively, then the prediction is considered correct. The threshold for experiments involving m=5 is set to three correct responses among the five meta-questions. More generally, in order to conclude the presence or absence of a particular category, the VQA model should be able to correctly answer at least (m+1)/2 meta-questions that are created based on the concept descriptions provided by GPT-3 for the respective object category. 

\section{Results \& Analysis over CUB Dataset}
\label{sec:res}

In this section, we report results over the CUB dataset  \cite{wah2011caltech} and perform detailed analysis of the proposed method including quality of GPT-3 \cite{brown2020language} generated concept descriptions, capability of existing VQA models to recognize fine-grained details in the images and comparison with existing concept learning methods (including zero-shot and one-shot approaches).
 
\paragraph{Zero-shot Performance}

We prepare variations of the zero-shot concept recognition approach described above by using different VQA systems (BLIP, GIT, and ViLT) and experiment with a varied number of GPT-3 generated concepts (i.e. when m is 1/3/5). We compare our approach with \citet{menon2022visual}, (which is referred as Ext-CLIP) which is the only method that supports zero-shot evaluation and reports results over CUB \cite{wah2011caltech} dataset. The performance with respect to the test partition of the CUB dataset is reported in Table \ref{tab:zsperf1}. Additionally, there is a column corresponding to m=0 in the Table \ref{tab:zsperf1}, which denotes the scenario where model does not take any LLM-generated concept descriptions into account. In other words, it can be thought of as having only a VQA model that takes test image and original test question (i.e. Is this a $<$category\_name$>$?). This is intended to serve as a VQA baseline without any concept learning procedure. 

\begin{table}[ht!]
\centering
\resizebox{\columnwidth}{!}{
\begin{tabular}{@{}llllll@{}}
\toprule 
& \multicolumn{5}{l}{\textbf{Accuracy(\%) with varied value of m}} \\
\multicolumn{1}{c}{\textbf{Method}}                                                                      & \multicolumn{1}{c}{0} & \multicolumn{1}{c}{1} & \multicolumn{1}{c}{3} & \multicolumn{1}{c}{5} & \multicolumn{1}{c}{var}                                  \\ \cmidrule(l){2-6} 
\multicolumn{1}{l|}{Ours- BLIP}                                                                                        & 54.30                 & \textbf{70.98}        & 62.15                 & 60.54                 & -                                                                     \\
\multicolumn{1}{l|}{Ours - GIT}                                                                                        & 45.86                 & 66.27                 &          58.34             & 56.78                 & -                                                                     \\
\multicolumn{1}{l|}{Ours- ViLT}                                                                                        & 48.33                 & 67.12                 &             59.05          &   53.28                    & -                                             \\ \cmidrule(l){1-6}                        
\multicolumn{1}{l}{\begin{tabular}[c]{@{}l@{}}Ext- CLIP\end{tabular}} & -                     & -                     & -                     & -                     & \begin{tabular}[c]{@{}l@{}}65.25 \end{tabular} \\ \bottomrule
\end{tabular}}
\caption{Performance comparison between variants of our zero-shot concept recognition approach and existing work by \citet{menon2022visual} denoted as Ext-CLIP over the CUB test-set. Accuracy (\%) is used as an evaluation metric as the underlying task is classification. `var' denotes the use of arbitrary number of concept descriptions produced by GPT-3 in the respective method.}
\label{tab:zsperf1}
\end{table}

 All of our zero-shot model variations with m=1 surpass the performance of Ext-CLIP. There are three distinctions between our model and Ext-CLIP; First, both approaches use slightly different prompts to obtain concept descriptions from GPT-3. We use `Describe in m phrases separated by \# -- how the $<$category\_name$>$ looks like' whereas Ext-CLIP uses `What are useful visual features for distinguishing a $<$category\_name$>$ in a photo?' prompt. Second, for a particular experiment, the value of m is pre-determined in our approach as the value of this parameter is used in our prompt template. Whereas VCD \cite{menon2022visual} takes into account an arbitrary number of concept descriptions for each object category returned by GPT-3. Therefore the performance of Ext-CLIP is reported under the column `var'. Finally, Ext-CLIP estimates CLIP similarity \cite{radford2021learning} between images and sentence `$<$category\_name$>$ which is/has $<$concept\_description$>$' at the test time. Contrary, we create a set of binary meta-questions based on the concept descriptions which do not directly include $<$category\_name$>$.

\begin{table}[b!]
$\frac{}{} \\$
\centering
\begin{tabular}{@{}lc@{}}
\toprule
\multicolumn{1}{c}{\textbf{Method}}   & \multicolumn{1}{l}{\textbf{Accuracy(\%)}} \\ \midrule
Ours- BLIP$_{one-shot}$ (m=1)       & \textbf{76.12}                            \\
Ours- BLIP$_{one-shot}$ (m=3)       & 69.46$^*$                            \\
Ours- BLIP$_{one-shot}$ (m=5)       & 63.22$^*$                            \\
Ext- Falcon                  & 75.28                            \\
Ext- LaBo (m = var)                   & 54.19                            \\
Ext- COMET (m<=15) & 67.9                             \\ \bottomrule
\end{tabular}
\caption{Accuracy(\%) comparison between a variant of our one-shot concept recognition approach and existing work by \citet{mei2021falcon, yang2023language, cao2020concept} denoted as Ext- Falcon, Ext- LaBo, Ext- COMET respectively over the CUB test-set (* indicates averaged results over three runs)}
\label{tab:osperf}
\end{table}
 
\paragraph{One-shot Performance}

There exists several methods in the literature \cite{mei2021falcon, yang2023language, cao2020concept} which are designed to learn new concepts from only one labelled example i.e. one-shot concept learning paradigm. Although our method is designed primarily for the zero-shot concept learning, to compare the performance with existing methods, we setup the one-shot model as follows:  As evident from the Table \ref{tab:zsperf1}, our proposed 4-step approach with BLIP as a VQA model performed the best. Therefore, we fine-tune BLIP with the test image and one meta-question for each CUB category. In case the exists more than one meta-questions (m=3 and m=5 cases), we randomly select one meta-question and use it in the training. We leverage this one-shot fine-tuned BLIP model in the Step 3 of the Section \ref{sec:llmvqa}. All the remaining steps are identical to the zero-shot pipeline. 

The results are summarized in Table \ref{tab:osperf} which compares our model with Ext- Falcon \cite{mei2021falcon}, Ext- LaBo \cite{yang2023language}, and Ext- COMET \cite{cao2020concept}. Ext- LaBo \cite{yang2023language} considers variable value of m in their model design (which are LLM generated concept descriptions) and Ext- COMET \cite{cao2020concept} considers at most 15 concepts (which are human-curated for ground-truth) per object category (i.e. m<=15). Ext- Falcon \cite{mei2021falcon} does not directly involve parameter m as their method attempts to learn visual concepts into box embeddings space. Since we randomly sample meta-questions in case of m>1, the reported results are averaged over three runs with a fixed seed value.

Our best model BLIP$_{one-shot}$ (m=1) yields the best accuracy of 76.12\%, which is 5.14\% improvement over its zero-shot counterpart (reported in Table \ref{tab:zsperf1}). The second best model is Ext- Falcon \cite{mei2021falcon}, tailing our best performing model by <1\% accuracy. The accuracy of our BLIP$_{one-shot}$ model variations decreased with the higher values of m (i.e. from 1 to 3 and 3 to 5), which is consistent to our observation in the zero-shot setting. Another interesting observation is that- for m=3, the performance gap between the zero-shot and one-shot model is maximum i.e. 7.31\%. This indicates a possibility of generated meta-questions (for m=3 case) being more useful for the BLIP model to solve the downstream concept identification task (compared to m=1 and m=5). Although more experiments are needed to make appropriate conclusion, this is an interesting research question to explore in future.

Other existing models (Ext- LaBo and Ext- COMET), despite incorporating more number of LLM generated or human-curated concepts, underperformed compared to our m=1 and m=3 model variations. Note that, among the models compared in Table \ref{tab:osperf}, LaBo \cite{yang2023language} and our proposed model are only two human-interpretable concept learning methods. Even though the Ext- Falcon \cite{mei2021falcon} achieves comparable accuracy to our model, it lacks natural language explanations conveying model's decision of a particular object category.

\paragraph{Effect of the Choice of m on Performance}

The VQA task over the CUB dataset referred in this work is a binary VQA task i.e. whether a particular bird is present in the image or not. Therefore, the random chance accuracy for this task is 50\%. The results from Table \ref{tab:zsperf1} indicate that VQA models that are not previously trained on the CUB data and directly tested over the bird identification task (m=0) perform almost close to the random chance. By simply replacing the category name with a single conceptual description generated by GPT-3 (i.e. m=1 case), the models gain 15-20\% accuracy improvements. When the value of m increases further, a drop in the overall accuracy has been observed across all VQA variants. This trend is consistent with one-shot variants of our model. 

As shown in Figure \ref{fig:cubmvar}, when m=1, the GPT-3 generated descriptions are relatively long and complex. In such cases, we suspect that VQA models might not carefully verify all attributes listed in the question. As a result, it is hard to pinpoint due to which attribute in the image When the concept descriptions are focused and concise (with higher values of m), models observe a drop in overall performance. Hence, there is a trade-off between performance and model transparency with increase in m value. In other words, the lower the m, the better the accuracy but the lesser the transparency. 

 \begin{figure}[b!]
\includegraphics[width=\linewidth]{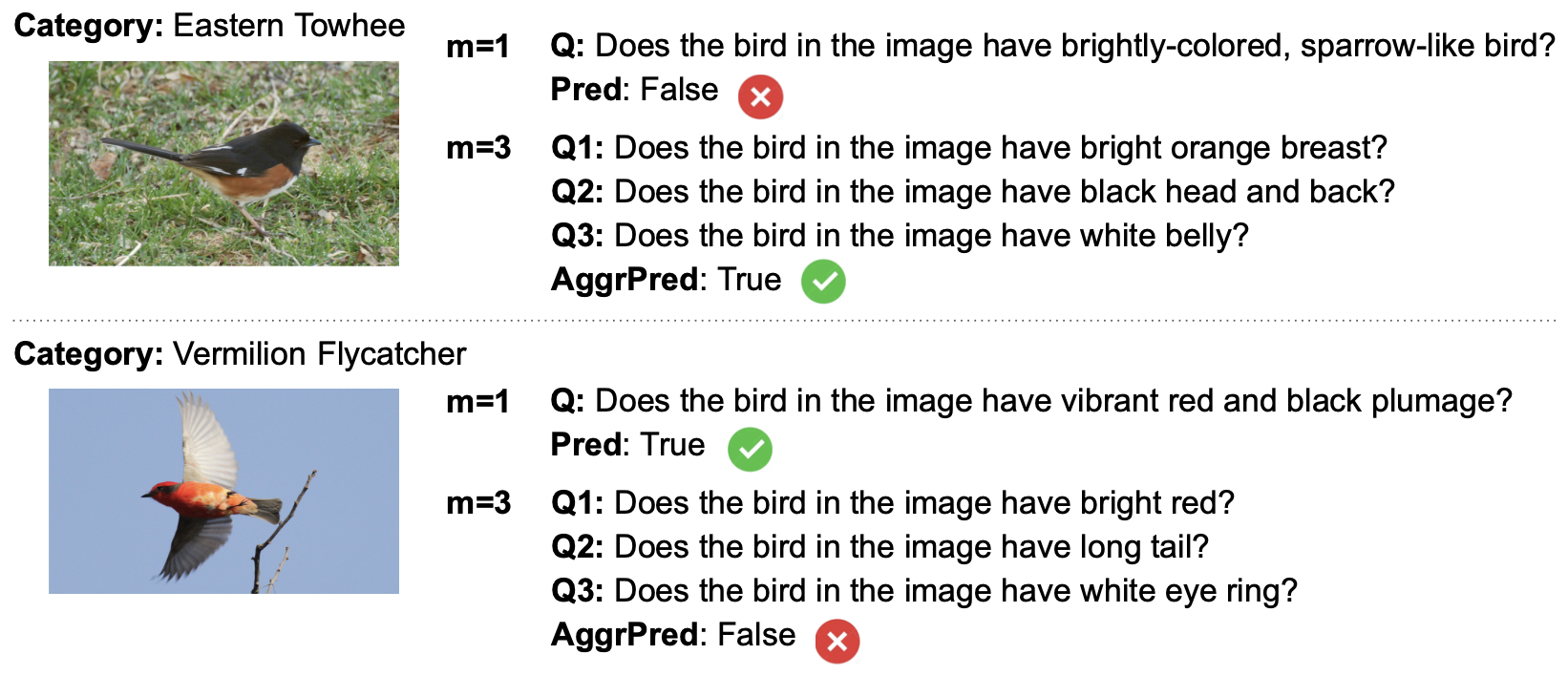}
\caption{Two qualitative examples predicted by the GPT-3+BLIP model, which is a top-performing zero-shot variant in our experiments.}
\label{fig:zsqual}
\end{figure}

\begin{figure}[ht!]
\centering
\includegraphics[width=\linewidth]{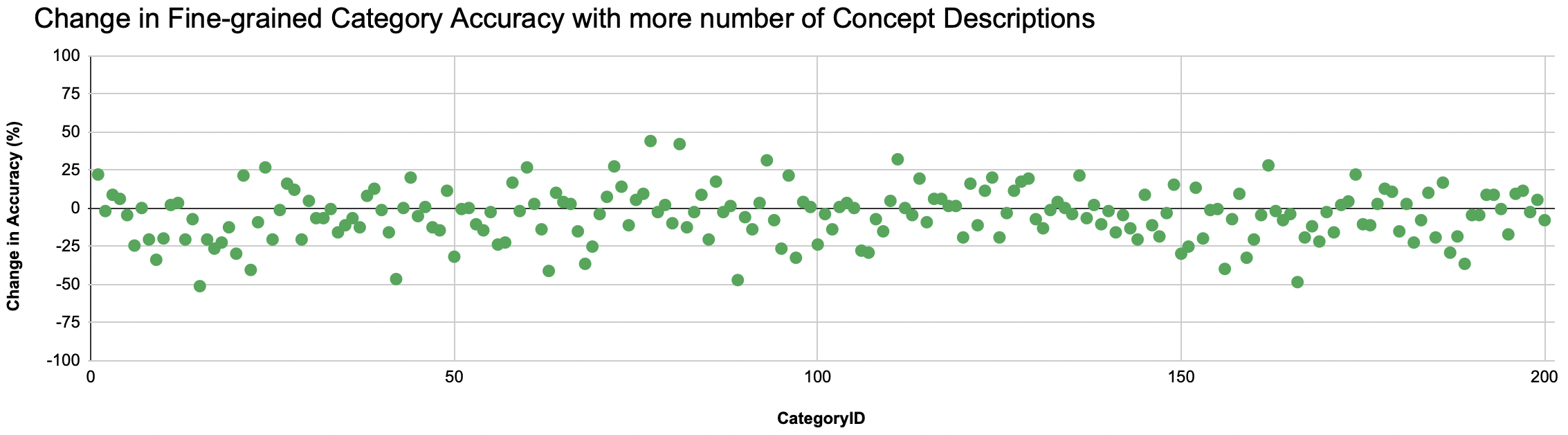}
\includegraphics[width=\linewidth]{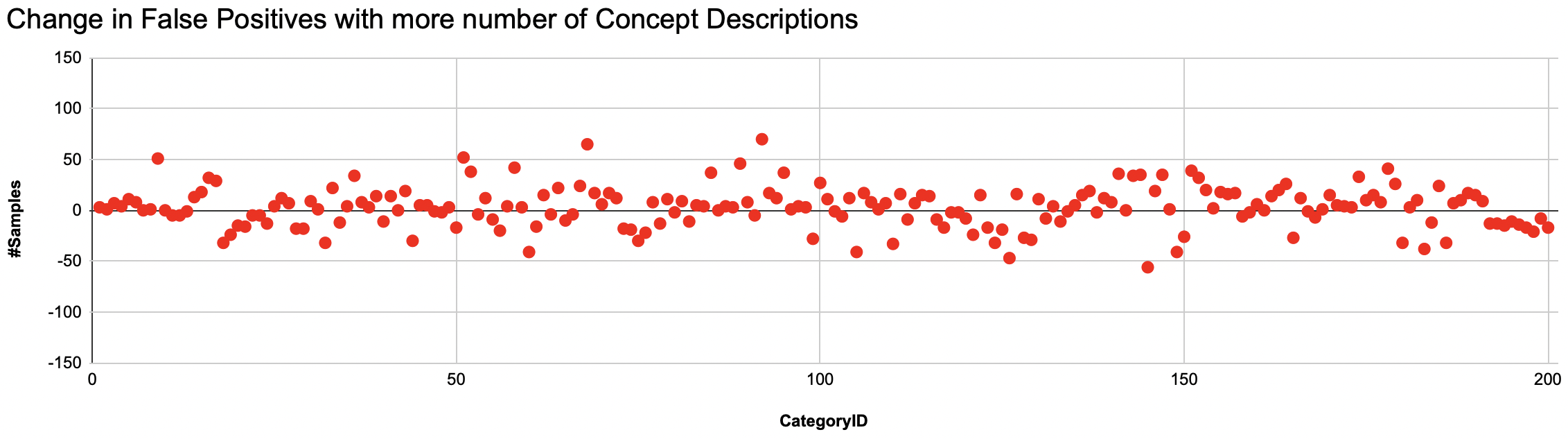}
\includegraphics[width=\linewidth]{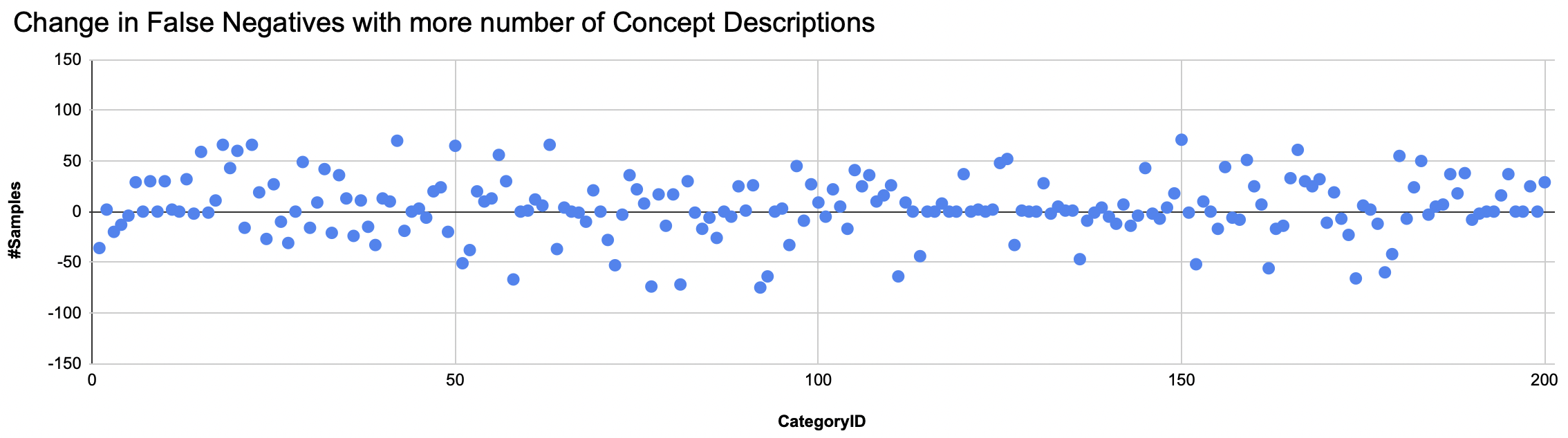}
\caption{Plots demonstrating how accuracy (on top), false positives (in mid), and false negatives (at the bottom) per object category change when more number of concept descriptions obtained using GPT-3 (i.e. when changing m=1 to m=3) are incorporated in our best zero-shot model GPT-3+BLIP.}
\label{fig:m-change}
\end{figure}

Two qualitative examples predicted by our best-performing zero-shot variant are shown in Figure \ref{fig:zsqual}. The example on the top demonstrates incorporating multiple concise concept descriptions is beneficial. Accuracy for `Eastern Towhee' category improves from 60.67\% to 82\% when m is set to 1 and 3 respectively. On the other hand, the example at the bottom demonstrates the case where a noisy visual attribute generated by GPT-3 (i.e. `white eye ring') hurts the model performance (when m=3). As a result, accuracy for `Vermilion Flycatcher' drops to 50.67\% (m=3) compared to 97.33\% (m=1). For both the success as well as the failure cases, our proposed model is human-interpretable.

We further analyzed how the choice of m affects the individual object categories. To visualize this, in Figure \ref{fig:m-change}, we plot how the accuracy, number of False Positives (FP), and False Negatives (FN) for each category varies when m changes from 1 to 3. Accuracy denotes the number of correct test predictions using our zero-shot LLM+BLIP variant. Among 200 object categories in the CUB dataset, 120 categories demonstrate lower overall accuracy, and for 74 categories the accuracy improves when m=3. Both accuracy gains and drops demonstrate 0-50\% variation in comparison with accuracy obtained for the m=1 setting. For 73 and 78 categories, the count of FN and FP predictions are respectively lower compared to m=1.





\paragraph{Diversity of LLM Generated Concept Descriptions}

\begin{figure}[ht!]
\includegraphics[width=\linewidth]{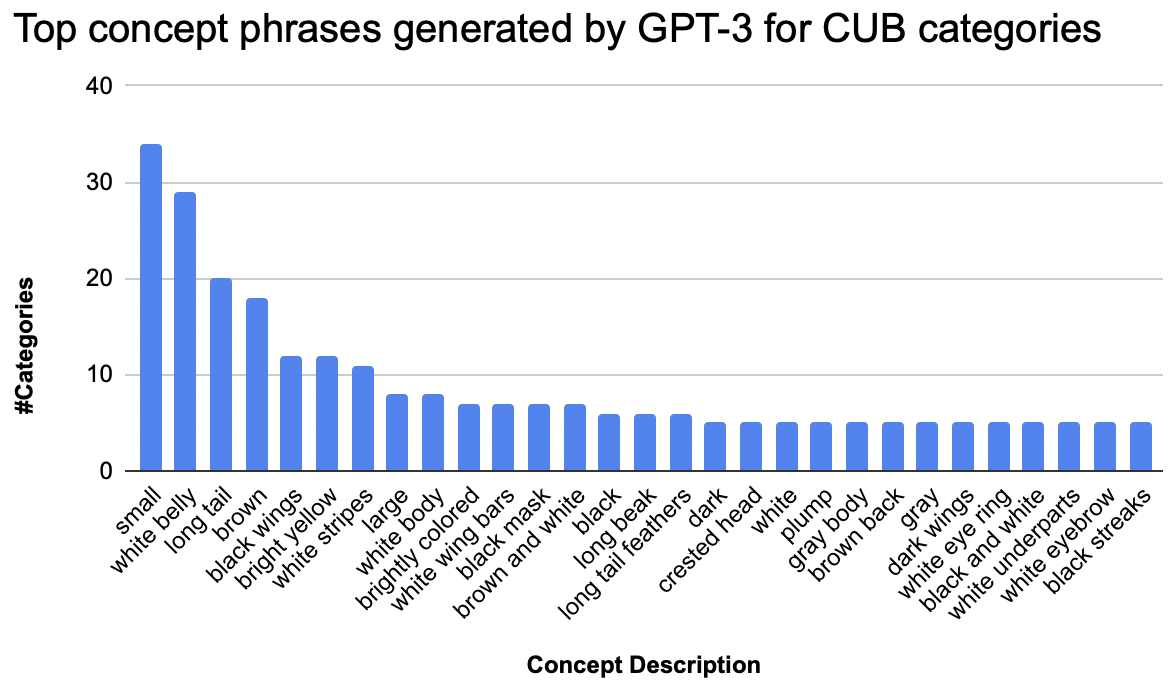}
\caption{Most frequent concept descriptions returned by GPT-3 for 200 object categories in the CUB dataset.$\frac{}{}\\$}
\label{fig:m-cub}
\end{figure} 

Here, we analyze how diverse GPT-3 generated concept descriptors are for all object categories in the CUB dataset. For 200 object categories, we analyzed top-3 concept descriptions (m=3) for each object category, among which 262 ($\sim$42\%) are unique. A plot of the most frequently generated descriptions for the CUB is shown in Figure \ref{fig:m-cub}. The phrases `small' and `white belly' are the most common across the dataset, specifically, shared across 34 and 29 categories respectively. 

\begin{figure}[ht!]
\centering
\includegraphics[width=0.8\linewidth]{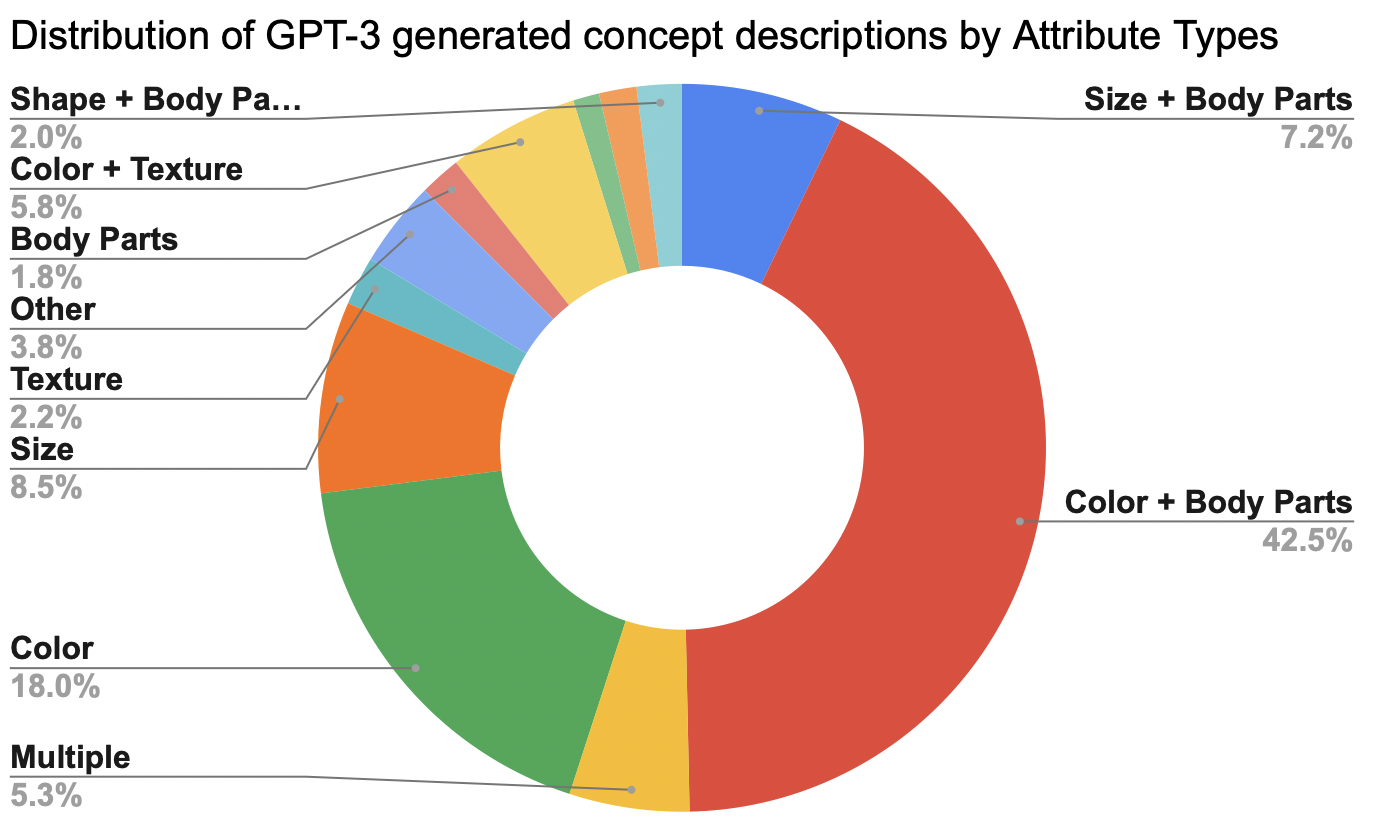}
\caption{Distribution of GPT-3 generated concept descriptions (m=3) for the CUB dataset as combinations of attributes (color, shape, size, texture/pattern, and body parts) commonly used to identify bird species.}
\label{fig:conceptdist}
\end{figure}

We observed that most concept descriptions generated by GPT-3 in the context of the CUB dataset include combinations of attributes such as color, shape, size, texture/pattern, and body parts. Therefore, we manually classified 600 concept descriptions based on the combination of attributes reflected. For example, the description `yellowish-olive' would be considered as `Color', `long tail' would be considered as `Size+Body Part', and so on. The distribution based on The label `Multiple' implies the presence of more than two attributes (e.g. `brown streaked back' which includes color, texture/pattern, and body parts), The label `Other' implies descriptions that do not fall into any of the standard combinations of attributes (e.g. `three toes on each foot' was a concept description obtained for the category american three toed woodpecker). The distribution of concept descriptions based on the attribute types is shown in Figure \ref{fig:conceptdist}. Among 600 descriptions analyzed, the attribute combinations `Color+Body Parts' (e.g. `yellow breast') and `Color' (e.g. `blue and white') cumulatively span 60\% of the descriptions.

\paragraph{Are VQA Systems Reliable for Fine-grained Attribute Recognition Described in Language?} 

Our proposed approach has two key components- an LLM and a VQA model. Hence, the end-task accuracy depends on- how correct are the LLM generated concept descriptions for various categories and how well the VQA system can verify the presence of those concept descriptions in the images. Assuming that GPT-3 has the perfect knowledge about appearance of various categories in the CUB dataset, in this section, we analyze how reliable are existing VQA models for the task of fine-grained attribute recognition. We previously identified five major attribute types i.e. colors, shapes, sizes, textures/patterns, body parts that are often used to describe CUB categories. For each attribute type, we randomly sample 3 examples from the dataset and show them in Table \ref{tab:qualex1}. 

We use two accuracy metrics- CategoryAccuracy and AttributeAccuracy for this evaluation. CategoryAccuracy denotes the correct number of answers generated by our proposed 4-stage model (GPT3+BLIP model with m=3) for all image-question pairs related to the given category in test set. On the other hand, AttributeAccuracy denotes the number of instances (within the test set for a given category) where the meta-question corresponding to the concept at hand (i.e. color `yellowish-olive' or texture `white spots on wings') is correctly verified by the model. Consider the example of `western grebe' category tested for VQA model's ability to recognize `Size' attribute (third row, first column in Table \ref{tab:qualex1}). The interpretation of CategoryAccuracy and AttributeAccuracy for this example can be justified as follows. Among all instances that aim to test `western grebe' category in CUB's test data, 84.67\% (i.e. CategoryAccuracy) of them can be correctly predicted by our model. On the other hand, for all those examples, the BLIP model can answer the meta-question `Does the bird in the image have long neck?' with accuracy 76\% (i.e. AttributeAccuracy).

\begin{table}[ht!]
\centering
\resizebox{\linewidth}{!}{%
\begin{tabular}{@{}clll@{}}
\toprule

\multicolumn{4}{c}{\textbf{Qualitative analysis of VQA system's ability to recognize various visual attribute types in CUB dataset}}                                                                             \\  \midrule
\multicolumn{4}{c}{\textbf{Color}} \\ \\                                                        
\multicolumn{1}{l}{\textbf{\begin{tabular}[c]{@{}l@{}}Image: \\ \\ \\ \\ \\ \\ \\ \\ Category: \\ Concept:\\ CategoryAcc(\%):\\ AttributeAccuracy(\%):\end{tabular}}} & \begin{tabular}[c]{@{}l@{}} \includegraphics[width=3.5cm,height=3.5cm]{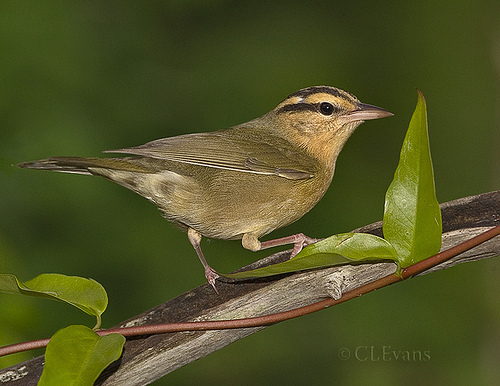} \\ worm eating warbler \\ yellowish-olive\\ 70.00\\ 97.33\end{tabular}     & \begin{tabular}[c]{@{}l@{}} \includegraphics[width=3.5cm,height=3.5cm]{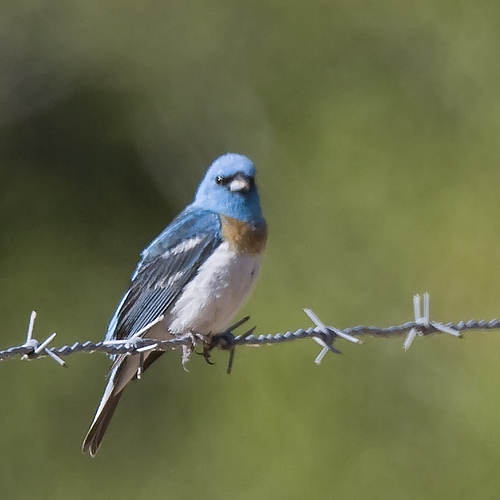} \\ lazuli bunting	\\ bright blue\\ 44.00\\ 97.33\end{tabular}                            & \begin{tabular}[c]{@{}l@{}} \includegraphics[width=3.5cm,height=3.5cm]{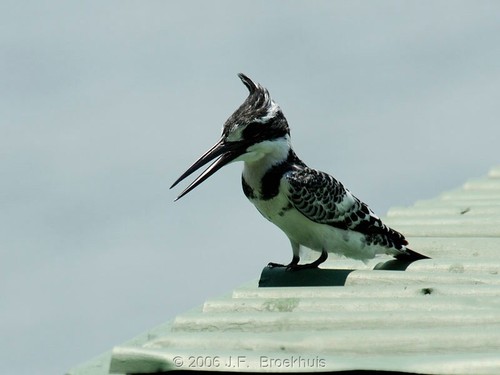} \\ pied kingfisher \\ blue and white	\\ 73.33\\ 30.67\end{tabular}      \\ \midrule  
\multicolumn{4}{c}{\textbf{Shape}} \\  \\                                                                
\multicolumn{1}{l}{\textbf{\begin{tabular}[c]{@{}l@{}}Image: \\ \\ \\ \\ \\ \\ \\ \\ Category: \\ Concept:\\ CategoryAcc(\%):\\ AttributeAccuracy(\%):\end{tabular}}} & \begin{tabular}[c]{@{}l@{}} \includegraphics[width=3.5cm,height=3.5cm]{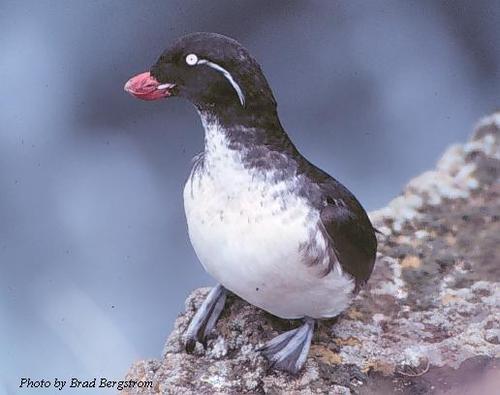} \\ parakeet auklet\\ round body \\ 49.33\\ 20.00\end{tabular}              & \begin{tabular}[c]{@{}l@{}} \includegraphics[width=3.5cm,height=3.5cm]{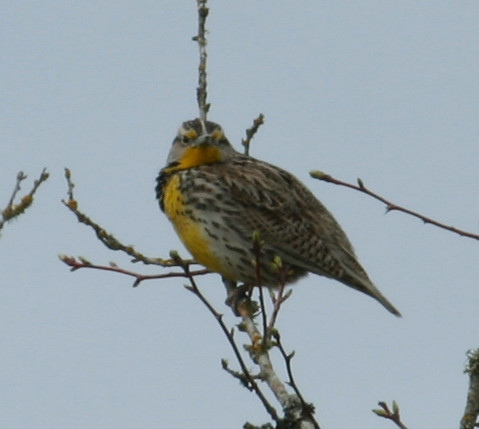} \\ western meadowlark \\ black v on chest\\ 52.67\\ 2.67\end{tabular}                     & \begin{tabular}[c]{@{}l@{}} \includegraphics[width=3.5cm,height=3.5cm]{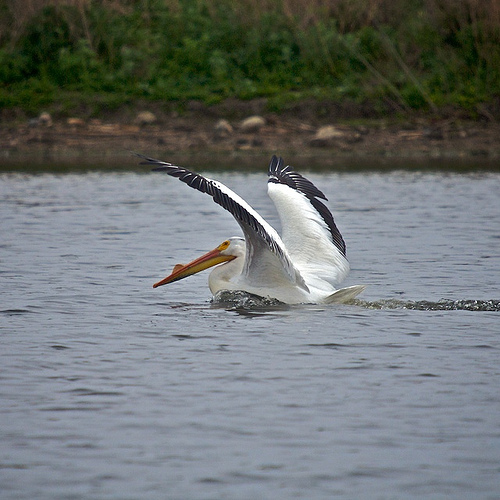} \\ white pelican \\ pouch-like bill\\ 91.33\\ 17.33\end{tabular}         \\ \midrule  
\multicolumn{4}{c}{\textbf{Size}}                                                             \\ \\
\multicolumn{1}{l}{\textbf{\begin{tabular}[c]{@{}l@{}}Image: \\ \\ \\ \\ \\ \\ \\ \\ Category: \\ Concept:\\ CategoryAcc(\%):\\ AttributeAccuracy(\%):\end{tabular}}} & \begin{tabular}[c]{@{}l@{}} \includegraphics[width=3.5cm,height=3.5cm]{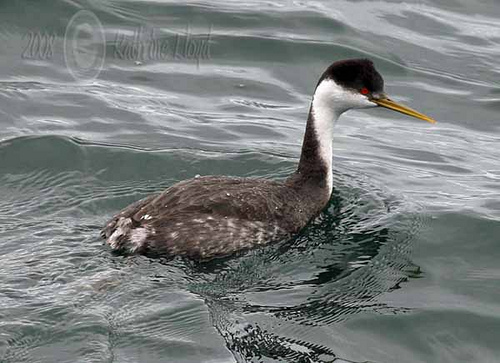} \\ western grebe \\ long neck\\ 84.67\\ 76.00\end{tabular}                 & \begin{tabular}[c]{@{}l@{}} \includegraphics[width=3.5cm,height=3.5cm]{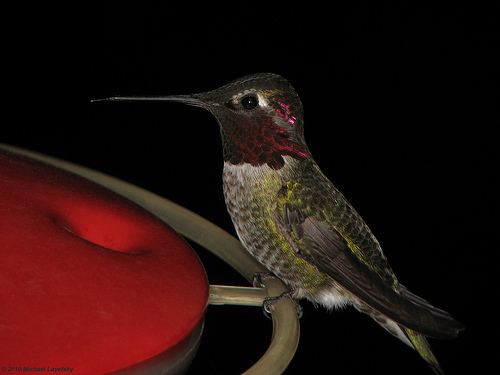} \\ anna hummingbird \\ long beak\\ 68.67\\ 81.33\end{tabular}                             & \begin{tabular}[c]{@{}l@{}} \includegraphics[width=3.5cm,height=3.5cm]{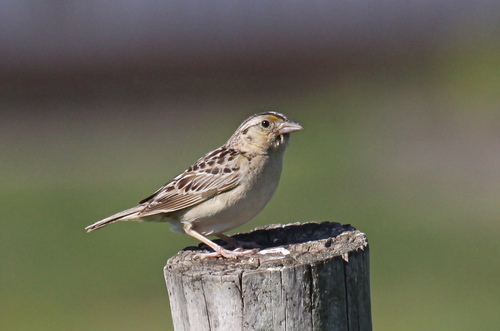} \\ grasshopper sparrow \\ small body\\ 71.33\\ 100.00\end{tabular}

\\  \midrule     
\multicolumn{4}{c}{\textbf{Texture}}                                                            \\  \\
\multicolumn{1}{l}{\textbf{\begin{tabular}[c]{@{}l@{}}Image: \\ \\ \\ \\ \\ \\ \\ \\ Category: \\ Concept:\\ CategoryAcc(\%):\\ AttributeAccuracy(\%):\end{tabular}}} & \begin{tabular}[c]{@{}l@{}} \includegraphics[width=3.5cm,height=3.5cm]{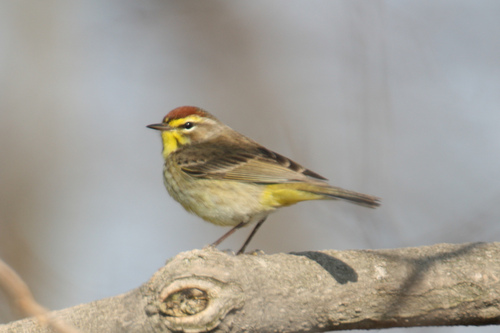} \\ palm warbler \\ streaked\\ 76.67\\ 80.00\end{tabular}                   & \begin{tabular}[c]{@{}l@{}} \includegraphics[width=3.5cm,height=3.5cm]{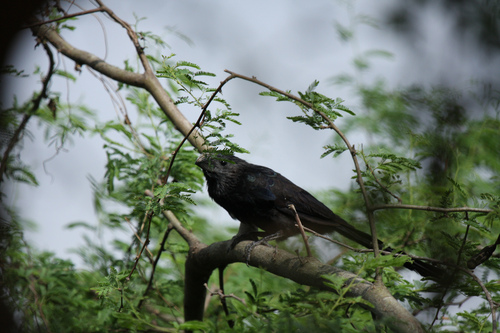} \\ groove billed ani \\ furry body\\ 43.33\\ 70.66\end{tabular} & \begin{tabular}[c]{@{}l@{}} \includegraphics[width=3.5cm,height=3.5cm]{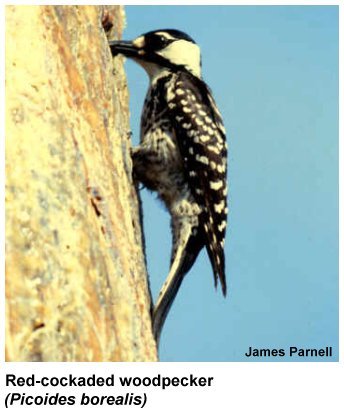} \\ red cockaded woodpecker\\ white spots on wings\\ 76.67\\ 85.33\end{tabular}              \\ \midrule       
\multicolumn{4}{c}{\textbf{Body Parts}}                                                              \\ \\
\multicolumn{1}{l}{\textbf{\begin{tabular}[c]{@{}l@{}}Image: \\ \\ \\ \\ \\ \\ \\ \\ Category: \\ Concept:\\ CategoryAcc(\%):\\ AttributeAccuracy(\%):\end{tabular}}} & \begin{tabular}[c]{@{}l@{}} \includegraphics[width=3.5cm,height=3.5cm]{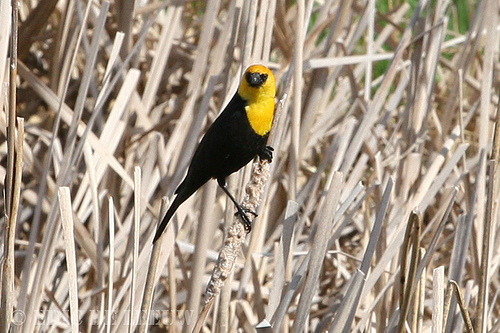} \\ yellow headed blackbird \\ white wing bars\\ 96.67\\ 18.67\end{tabular}  & \begin{tabular}[c]{@{}l@{}} \includegraphics[width=3.5cm,height=3.5cm]{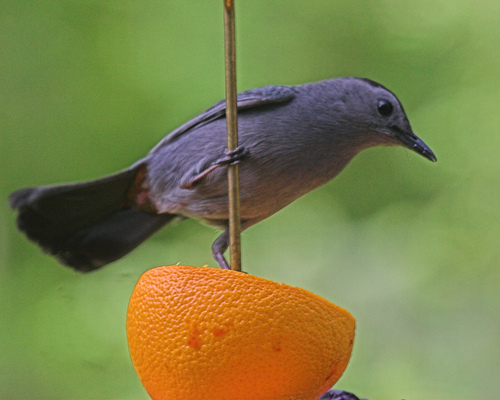} \\ gray catbird \\ white undertail\\ 38.00\\ 0.00\end{tabular} & \begin{tabular}[c]{@{}l@{}} \includegraphics[width=3.5cm,height=3.5cm]{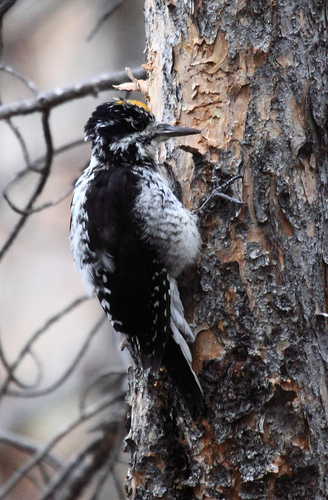} \\ american 3-toed woodpecker \\ three toes on each foot\\ 48.67\\ 18.67\end{tabular}  \\ \midrule         
\multicolumn{4}{c}{\textbf{Other Adjectives}}                                                     \\ \\
\multicolumn{1}{l}{\textbf{\begin{tabular}[c]{@{}l@{}}Image: \\ \\ \\ \\ \\ \\ \\ \\ Category: \\ Concept:\\ CategoryAcc(\%):\\ AttributeAccuracy(\%):\end{tabular}}} & \begin{tabular}[c]{@{}l@{}} \includegraphics[width=3.5cm,height=3.5cm]{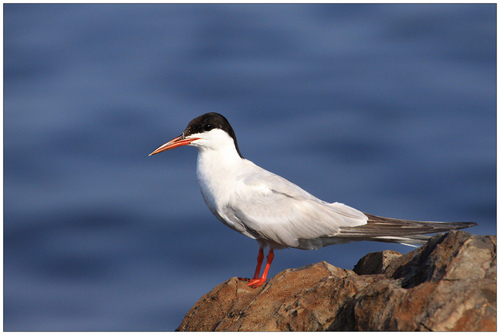} \\ common tern\\ long forked tail\\ 73.33\\ 30.67\end{tabular}            & \begin{tabular}[c]{@{}l@{}} \includegraphics[width=3.5cm,height=3.5cm]{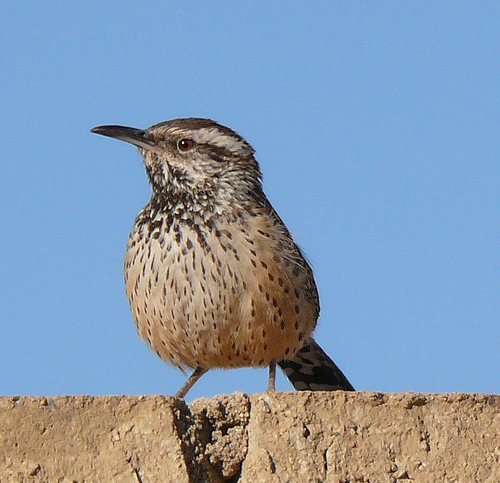} \\ cactus wren \\ spiky feathers\\ 78.00\\ 96.00\end{tabular}                             & \begin{tabular}[c]{@{}l@{}} \includegraphics[width=3.5cm,height=3.5cm]{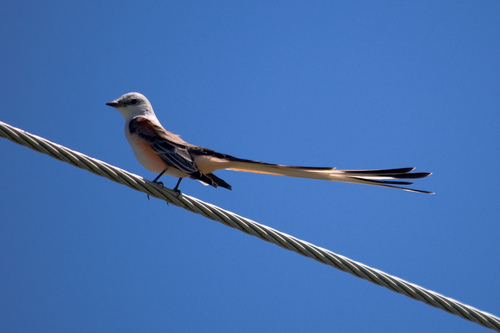} \\ scissor tailed flycatcher\\ bright colors\\ 60.67\\ 21.33\end{tabular} 
\\    \bottomrule
\end{tabular}*
}
\caption{Qualitative analysis of VQA system's ability to recognize color, shape, size, texture, body parts and other adjectives generated by GPT-3 for various CUB object categories.}
\label{tab:qualex1}
\end{table}

Observing the results in Table \ref{tab:qualex1}, we can conclude that attributes `Color', `Size', and `Texture' are robustly recognized by the BLIP model. Whereas the AttributeAccuracy remains low for `Shape' and `Body Parts', indicative of model struggling with these attributes despite achieving high CategoryAccuracy for certain examples. This demonstrates the advantage of using higher values of m (such as 3 or 5) in our proposed model. In other words, by incorporating more than one fine-grained concept descriptions, we can prevent single point of failures in the VQA stage of the model by relying on other visual attributes. This is especially the case where the generated concept description are hard to recognize in the image due to pose variations or details captured in the image.  For example, consider the `Scissor tailed flycatcher' category in Table \ref{tab:qualex1} (bottom right corner). The concept generated by GPT-3 for this category is `bright colors'. Due to lighting in this picture, it is hard to conclude whether this bird has bright colors or not. The BLIP model also struggles to recognize this attribute, evident from the low AttributeAccuracy i.e. 21.33\%. However, the respective CategoryAccuracy is relatively high (60.67\%), which indicates that model is possibly relying on other concept descriptions returned by GPT-3 and can correctly answer meta-questions based on them.

\section{Results for Other Visual Classification Datasets}


\begin{table}[ht!]
\centering
\begin{tabular}{@{}llll@{}}
\toprule
\multicolumn{1}{c}{\textbf{Dataset}} & \multicolumn{1}{c}{\#\textbf{Categories}} & \multicolumn{1}{c}{\textbf{Domain}} & \multicolumn{1}{c}{\textbf{Size}} \\ \midrule
CIFAR-10                & 10                                                                                   & Mixed                      & 10k                                \\
CIFAR-100                   & 100                                                                                  & Mixed                      & 10k                                \\
Food-101                    & 101                                                                                  & Food                       & 25.2k                              \\
DTD                         & 47                                                                                   & Textures                   & 1.8k                               \\ \bottomrule
\end{tabular}
\caption{{Image classification datasets used for evaluation in this paper along with relevant information about number of distinct categories incorporated, domain of the images and size of the testing data.$\frac{}{}\\$}}
\label{tab:datasets}
\end{table}

To verify the generalization ability of the proposed LLM+VQA pipeline for concept learning about objects, in this section, we conduct testing over more  image classification benchmarks. For this purpose, we leverage popular image classification datasets \cite{ krizhevsky2009learning, bossard2014food, cimpoi2014describing}. The characteristics of these datasets are briefly listed in Table \ref{tab:datasets}. We programmatically convert these datasets to the zero-shot VQA format described in Section \ref{sec:llmvqatask}. We compare our method against LaBo \cite{yang2023language}, CDA \cite{yan2023learning}, CuPL \cite{pratt2022does} and VCD \cite{menon2022visual}. Brief description of these methods was covered in Section \ref{sec:vclmlit}. Although there are many other image classifications datasets (not covered here), our choice of datasets is based on the popularity among the methods that we compare against. The only exception is the VCD model, which does not report performance on CIFAR variants.

The results are reported in Table \ref{tab:zsperf}. As explained earlier, the choice of parameters m, n and p play a crucial role in our proposed method. We report m, n and p parameters for each method (to the best of our knowledge and understanding of the architecture presented in the respective manuscript). Although due to the difference in these parameters, these methods cannot be compared directly, we suggest to use the information in Table \ref{tab:zsperf} to better understand the pros/cons of each method and their cumulative performance as a measure of how hard a particular image classification dataset is and to what extent LLMs are able to generate appropriate descriptions about the classes.

\begin{table*}
\resizebox{\linewidth}{!}{
\begin{tabular}{@{}llllllllllll@{}}
\toprule
                 &                                                                                                & \multicolumn{10}{c}{\textbf{Method (and corresponding Accuracy\%)}}                                                                                                                                                                                                                                                                                                                               \\ \midrule
\textbf{Dataset} & \multicolumn{1}{c}{\textbf{\begin{tabular}[c]{@{}c@{}}Configurable\\ Parameters\end{tabular}}} & \textbf{LaBo} & \textbf{CDA}        & \textbf{CuPL}    & \textbf{VCD}     & \begin{tabular}[c]{@{}l@{}}\textbf{Our}\\ \textbf{(BLIP)}\end{tabular} & \begin{tabular}[c]{@{}l@{}}\textbf{Our}\\ \textbf{(BLIP)}\end{tabular} & \begin{tabular}[c]{@{}l@{}}\textbf{Our}\\ \textbf{(BLIP)}\end{tabular} & \begin{tabular}[c]{@{}l@{}}\textbf{Our}\\ \textbf{(ViLT)}\end{tabular} & \begin{tabular}[c]{@{}l@{}}\textbf{Our}\\ \textbf{(ViLT)}\end{tabular} & \begin{tabular}[c]{@{}l@{}}\textbf{Our}\\ \textbf{(ViLT)}\end{tabular} \\ \midrule
                 & \textbf{m}                                                                                     & 50            & 4          & 1       & var     & 1                                                    & 3                                                    & 5                                                    & 1                                                    & 3                                                    & 5                                                    \\
                 & \textbf{n}                                                                                     & 1             & 0        & 0       & 0       & 0                                                    & 0                                                    & 0                                                    & 0                                                    & 0                                                    & 0                                                    \\
                 & \textbf{p}                                                                                     & fix           & fix        & var     & fix     & fix                                                  & fix                                                  & fix                                                  & fix                                                  & fix                                                  & fix                                                  \\
\textbf{CIFAR-10}         &                                                                                                & $\sim$92\%    & $\sim$67\% & 95.81\% & -       & 61.42\%                                              & 64.28\%                                              & 51.37\%                                              & 42.85\%                                              & 62.81\%                                              & 41.81\%                                              \\
\textbf{CIFAR-100}        &                                                                                                & $\sim$63\%    & $\sim$17\% & 78.47\% & -       & 26.06\%                                              & 33.73\%                                              & 22.29\%                                              & 27.11\%                                              & 31.21\%                                              & 16.27\%                                              \\
\textbf{Food-101}         &                                                                                                & $\sim$80\%    & $\sim$17\% & 93.33\% & 83.63\% & 82.95\%                                              & 86.27\%                                              & 84.30\%                                              & 79.37\%                                              & 79.77\%                                              & 80.02\%                                              \\
\textbf{DTD}              &                                                                                                & $\sim$53\%    & -          & 58.90\% & 44.26\% & 83.83\%                                              & 88.94\%                                              & 90.80\%                                              & 87.87\%                                              & 94.45\%                                              & 96.12\%                                              \\ \bottomrule
\end{tabular}}
\caption{{Performance comparison between variants of our zero-shot concept recognition approach and existing work related to concept learning using LLMs over multiple image classification datasets. Parameters n, m and p are reported for each method for comparison. Accuracy (\%) is used as an evaluation metric as the underlying task is classification. Approximate ($\sim$) symbol is used when exact value is not available. Non-availability of the respective result is indicated by (-) or unk.}}
\label{tab:zsperf}
\end{table*}

As observed in Table \ref{tab:zsperf}, most of the existing models for concept learning use fixed/templated prompt to obtain concept descriptions from the LLM. The only exception is the CuPL \cite{pratt2022does} model. Additionally, our method, CuPL and VCD \cite{menon2022visual} are designed for zero-shot concept recognition. Whereas LaBo \cite{yang2023language} requires at least 1 training example for each category. Finally, for our model, CDA and VCD rely on a limited set of concept descriptions retrieved with the help of LLM. Contrary to that, LaBo requires significantly large number of concept descriptions (50 descriptions per object category) to push the model performance. On the other hand, VCD dynamically determines the number of concept descriptions based on the example at hand (denoted as var).

The CuPL method achieves the best overall performance over CIFAR-10, CIFAR-100 and Food-101 datsets. All of our proposed variants outperform other methods over DTD dataset, with notable performance gains. GPT3+ViLT combination with 5 concept descriptions yields the best performance overall (96.12\%) for the DTD dataset. Moreover, the DTD dataset poses exception to our prior observation that performance boosts when changing m=1 to m=3 and declines when further increases to m=5. This possibly indicates that the concepts to be learnt (textures in this case) benefit from having multiple alternative descriptions that the model can rely upon and avoid failure from limited attributes that may or may not be recognizable in the images. Our proposed model demonstrates the second best performance over Food-101 dataset, tailing behind 7.06\% by CuPL. LaBo achieves the comparable performance over the CIFAR-10 dataset in comparison with the best method CuPL. However, with more number of classes in CIFAR-100, the performance gap between CuPL and LaBo increases. CDA consistently remains the low performing method across all datasets considered. The key takeaways from these experiments are summarized below:

\begin{itemize}[noitemsep, leftmargin=*]
    \item Concept learning methods that leverage LLMs work very well under zero-shot settings, which can be a promising alternative to expensive supervised training, which is expensive and time-consuming in nature.
    \item The construction of the prompt is a crucial step for all concept learning methods based on LLMs, The customized prompt formulation is likely to achieve superior results.
    \item In an ideal case, m>1 is desirable to avoid single point of failure in concept recognition pipeline, which is the unique aspect of our model proposed in this paper.
\end{itemize}

\section{Conclusion}

Humans capitalize on concepts to learn mental representations of various objects and use this knowledge to identify novel instances of an object using only a few examples. Inspired by this, in this work, we introduce a framework for zero-shot visual concept learning. We leverage the linguistic knowledge about visual objects from large language models (LLM) to generate concept descriptors. Instead of asking a VQA model whether or not a particular category is present in the image, we break down the category identification process into multiple sub-questions based on the LLM generated concept descriptors. Using GPT-3 along with the state-of-the-art VQA models, we demonstrate promising results over the existing fine-grained image recognition datasets. Our proposed method does not require significant computing overhead, improves performance over existing methods, and is human-interpretable to explain the model decisions.

\section*{Limitations}

In this paper, we collect textual descriptions of how various objects around us look like with the help of GPT-3 \cite{brown2020language} model. Therefore, the performance of our proposed approach relies on the quality and availability of necessary knowledge possessed by the large language model (LLM). This study was intended to gain insights on LLM's capability to learn concepts (without explicitly being trained on) through a combination of quantitative and qualitative metrics. From these results, it is still unclear for which kinds of datasets (coarse-grained or fine-grained) the model should be preferred over traditional concept learning methods or supervised image classification models. In this work, we limit our experiments to GPT-3 and demonstrate a proof of concept. However, there are several other LLMs that exist\footnote{https://github.com/Hannibal046/Awesome-LLM} and more advanced ones are coming out on a regular basis. Based on this work, we cannot advise on other LLMs ability to perform concept learning. Another downside of large language models (in general) is that their behaviour is hard to reproduce even with the identical prompt and use of parameters such as temperature and top-p. The cost involved with the proprietary LLMs is definitely a considerable factor, but we hope that in future, LLMs with less number of parameters, competitive performance and open-source variants would be available to the researchers and practitioners. We acknowledge this limitation and keep the floor open to for future investigations in this area. 


\section*{Ethics Statement}
In this paper, we work with pre-trained GPT3 model \cite{brown2020language} to generate textual descriptions of how various objects around us look like. As our efforts were limited to the publicly available image classification datasets (which includes birds, everyday objects, textures and food items), there was no content generated with offensive language or known discrimination concerning racial/gender aspects. However, several challenges concerning GPT-3 (including hallucinations and harmful content generation) are well-known and text generation results might be affected depending on the input prompts. 


\bibliography{emnlp2023}
\bibliographystyle{acl_natbib}

\appendix



\end{document}